\documentclass[final]{cvpr}

\usepackage{times}
\usepackage{epsfig}
\usepackage{graphicx}
\usepackage{amsmath}
\usepackage{amssymb}

\newcommand{\tabincell}[2]{\begin{tabular}{@{}#1@{}}#2\end{tabular}}
\usepackage{multirow}
\usepackage[caption=false]{subfig}
\DeclareMathOperator*{\argmax}{arg\,max}

\usepackage{algorithmic}
\usepackage[ruled]{algorithm2e}
\usepackage{calc}
\usepackage{enumitem}
\usepackage{paralist}

\SetKwInput{KwInput}{Input}
\SetKwInput{KwOutput}{Output}

\usepackage[pagebackref=true,breaklinks=true,colorlinks,bookmarks=false]{hyperref}
\def\cvprPaperID{10200} 
\def\confYear{CVPR 2021}

\begin{document}

\title{Robust and Accurate Object Detection via Adversarial Learning 
}

\author{
Xiangning Chen\textsuperscript{1,2}\thanks{Work done during an internship at Google.}
\enskip Cihang Xie\textsuperscript{3}
\enskip Mingxing Tan\textsuperscript{1}
\enskip Li Zhang\textsuperscript{1}
\enskip Cho-Jui Hsieh\textsuperscript{2}
\enskip Boqing Gong\textsuperscript{1}\\ \vspace{-0.3em}
\\
\textsuperscript{1}Google \enskip\enskip\enskip\enskip\enskip\enskip \textsuperscript{2}UCLA \enskip\enskip\enskip\enskip\enskip\enskip \textsuperscript{3}UCSC\\
}

\maketitle

\begin{abstract}

Data augmentation has become a de facto component for training high-performance deep image classifiers, but its potential is under-explored for object detection. Noting that most state-of-the-art object detectors benefit from fine-tuning a pre-trained classifier, we first study how the classifiers' gains from various data augmentations transfer to object detection. The results are discouraging; the gains diminish after fine-tuning in terms of either accuracy or robustness. This work instead augments the fine-tuning stage for object detectors by exploring adversarial examples, which can be viewed as a model-dependent data augmentation. Our method dynamically selects the stronger adversarial images sourced from a detector's classification and localization branches and evolves with the detector to ensure the augmentation policy stays current and relevant. This model-dependent augmentation generalizes to different object detectors better than AutoAugment, a model-agnostic augmentation policy searched based on one particular detector. Our approach boosts the performance of state-of-the-art EfficientDets by +1.1 mAP on the COCO object detection benchmark. It also improves the detectors' robustness against natural distortions by +3.8 mAP and against domain shift by +1.3 mAP. Models are available at \url{https://github.com/google/automl/tree/master/efficientdet/Det-AdvProp.md}.

\end{abstract}

\begin{figure}
\centering
\includegraphics[width=.9\columnwidth]{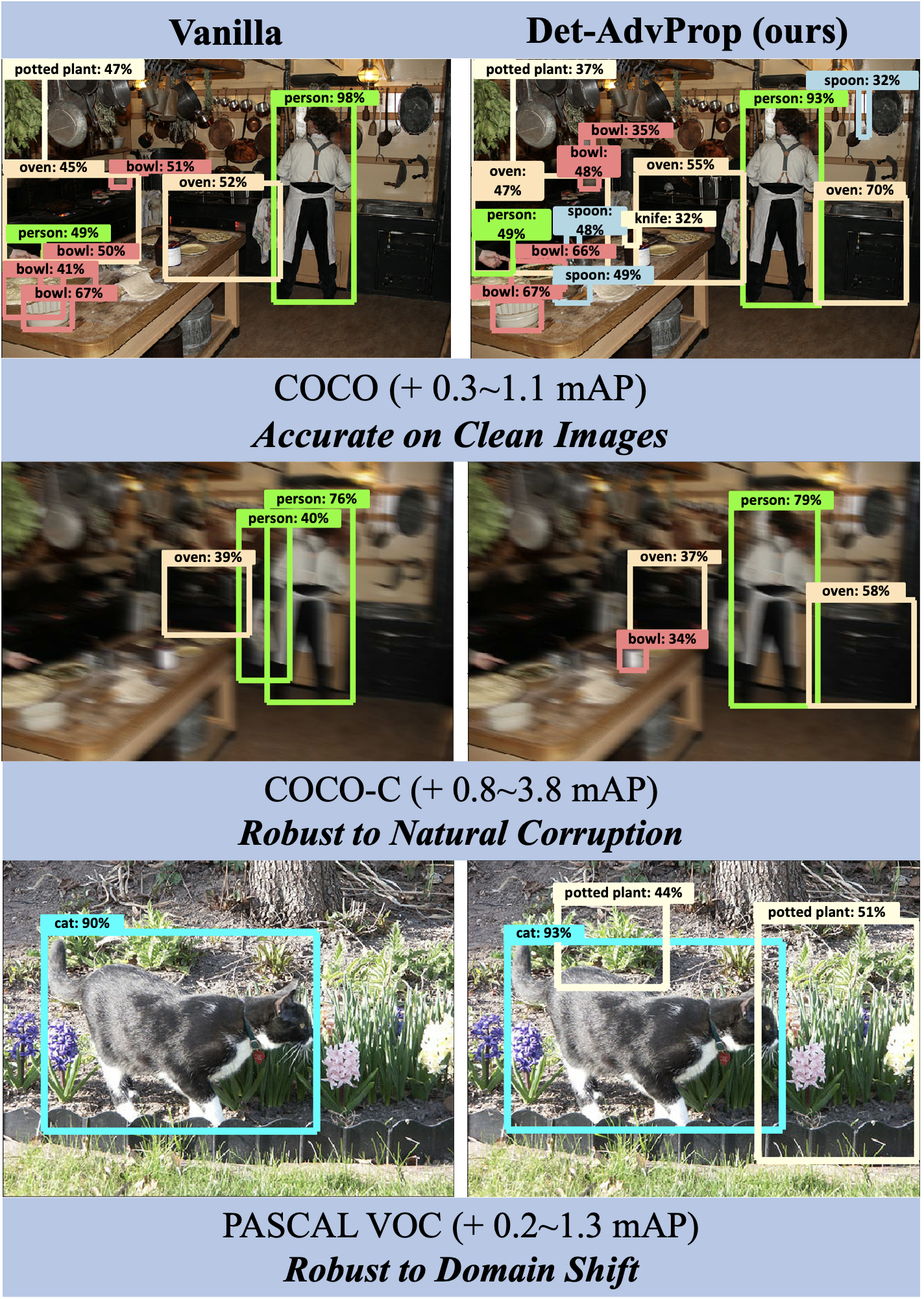}
\caption{
\textbf{Top:}
Det-AdvProp improves object detectors' accuracy on clean images.
Our model correctly detects some objects (e.g., ``spoon'' and ``knife'') missed by the vanilla detector trained without Det-AdvProp.
\textbf{Middle:}
Det-AdvProp improves the detectors' robustness against natural corruption.
The vanilla detector misses ``bowl'' and ``oven'' and produces a false positive for ``person'' after the image is corrupted by motion blur.
\textbf{Bottom:}
Det-AdvProp improves robustness against cross-dataset domain shift.
We can successfully detect the ``potted plants'' behind the ``cat'' from an image out of the dataset for training. (best viewed in color).
}
\vspace{-12pt}
\label{fig:teaser}
\end{figure}

\section{Introduction}
Deep neural networks (DNNs) are powerful tools for visual representation learning. As the training data grows in size and diversity, DNNs keep up the pace and achieve unprecedented performance on a wide range of benchmarked tasks~\cite{he2016resnet, ren2015faster, redmon2016yolo, long2015fcn, he2017mask, qi2018frustum}. The learned representations also demonstrate good transferability to downstream tasks for which there is often a small amount of curated data. This pre-training and then fine-tuning paradigm is one of the crucial enablers for state-of-the-art object detectors~\cite{tan2019enet, he2019rethinking, redmon2016yolo, lin2017focal, liu2016ssd}. In this paper, we aim to enhance this learning paradigm for training not only accurate but also robust object detectors.


We first revisit the role of pre-training in object detection, given He \etal.'s study~\cite{he2019rethinking} about vanilla ImageNet~\cite{deng2009imagenet} pre-training and yet the new advances in data-augmented ImageNet pre-training~\cite{xie2020self, xie2020advprop}. 
We examine both the accuracy and robustness of the detectors. 
In the study with the top-performing EfficientDet detectors~\cite{tan2020edet}, we find that the performance gains for ImageNet classification, brought by advanced data augmentation methods in pre-training, diminish after fine-tuning regardless of the detectors' accuracy or robustness. 
This observation motivates us instead to investigate fine-tuning, the second stage in the paradigm for training object detectors.



Our high-level idea is to reposition the recently developed data augmentations from the pre-training stage to fine-tuning. We first study AutoAugment~\cite{zoph2019learning} in the object detector fine-tuning because its policy is purposely searched for augmenting object detectors' training. However, experiments reveal that AutoAugment fails to provide consistent improvements to the detectors we studied, probably because it was searched based on only one object detector and one dataset, limiting its generalization ability. 


In light of the lessons above, we switch to the model-dependent data augmentation of AdvProp~\cite{xie2020advprop} for fine-tuning object detectors. AdvProp uses adversarial examples to improve image classification models. It employs separate batch normalization layers for the clean training images and the adversarial examples to accommodate their distinct statistics. Unlike AutoAugment or many other augmentation methods, AdvProp can dynamically evolve with the primary model during training to ensure the augmentation is up to date. 

We improve AdvProp to fit it into the object detection fine-tuning (denoted by Det-AdvProp). 
Previous works show that detectors benefit from shape cues~\cite{geirhos2018imagenettrained}, and adversarial examples help CNNs learn shape-related representations~\cite{zhang2019interpreting}.
There are two sources to generate adversarial examples using a detector: its classification head and its localization head. We conduct a local comparison at each training iteration to identify the source that is more ``adversarial'' than the other, which is then selected to augment the training data.
We show that this local comparison is crucial. Straightforwardly aggregating the two sources gives rise to weak adversarial examples because some of the adversarial gradients mutually conflict~\cite{zhang2019towards}. 
Another alternative, keeping both and separating them to different batch normalization layers, incurs a too strong regularization to the detector, leading to low accuracy, albeit high robustness. 

We report the following main findings in this paper. 
Although the pre-training stage remains more effective for object detection than random initialization under a reasonable computing budget (we run the fine-tuning for up to 300 epochs on the COCO object detection dataset~\cite{lin2015coco}),
the performance gain at the pre-training-stage diminishes after fine-tuning, regardless of any strong data augmentations for the pre-trained backbone. Instead, we demonstrate that it is more promising to incorporate advanced data augmentations into fine-tuning. Our Det-AdvProp boosts state-of-the-art EfficientDets' accuracy by 0.3--1.1 mAP, robustness to natural corruption by 0.8--3.8 mAP, and robustness to domain shift by 0.2--1.3 mAP (illustrated in Figure~\ref{fig:teaser}). 
Finally, we see that our model-dependent Det-AdvProp substantially outperforms the model-agnostic AutoAugment in object detection under various settings.




\section{Related Work}

\paragraph{Data Augmentation.}
By applying label-preserving transformation to images, data augmentation has become a standard paradigm for training an image classifier~\cite{cubuk2019autoaug, cubuk2020randaug, xie2020advprop, zhang2018mixup, lemley2017smart, lim2019fast, zhang2020adversarial, li2020shapetexture}.
Most of them are model agnostic policy.
In comparison, AdvProp~\cite{xie2020advprop} augments the training data by adversarial attack, its augmentation policy is unique to model and data.
For the downstream object detection task, there are much fewer works specifically crafted for the fine-tuning process~\cite{zhang2019bag, zoph2019learning}.
One representative work - AutoAugement~\cite{zoph2019learning} searches based on RetinaNet~\cite{lin2017focal} and a subset of COCO dataset~\cite{lin2015coco}, containing operations like rotation and bbox-only-translation.


\vspace{-10pt}
\paragraph{Attacks and Adversarial Training for Object Detectors.}
Many effective attacks crafted for object detectors have been proposed recently. 
Most of them generate adversarial examples solely based on one individual loss (classification or localization loss) of the detection task~\cite{xie2017adversarial, chen2019shapeshifter, kevin2018phyattack, lu2017adversarial}. 
For instance, 
DAG~\cite{xie2017adversarial} optimizes over a loss function that misleads the detectors to produce incorrect classification results.
Some other works~\cite{li2019exploring, liu2019dpatch} simultaneously attacks both the bounding box regression and classification to disable their predictions.
To defend those attacks, Zhang \etal.~\cite{zhang2019towards} extend adversarial training~\cite{madry2018towards} to the scenario of object detection by leveraging the attacks sourced from both classification and localization domains. 
Our proposed Det-AdvProp can largely outperform their adversarial training in terms of both clean accuracy and robustness (Appendix~\ref{app:adv}).



\section{Motivation}
Data augmentation is an effective way to improve image classification models~\cite{cubuk2019autoaug, zhang2018mixup, cubuk2020randaug, xie2020self}.
For example, AdvProp~\cite{xie2020advprop} uses adversarial images to boost the accuracy of EfficientNets~\cite{tan2019enet} by up to 0.6\% on ImageNet~\cite{deng2009imagenet} accuracy, and Noisy Student~\cite{xie2020self} trained with noise and data as augmentation surpasses the vanilla models by over 1\% on ImageNet. When it comes to object detection, which is arguably more complex than image classification, there are two natural choices to augment a detector:
(i) Borrow an already augmented model from the upstream classification task and then fine-tune it for object detection. 
(ii) Directly augment the detector during training.

\begin{figure}[!htb]
\centering
\includegraphics[width=0.75\columnwidth]{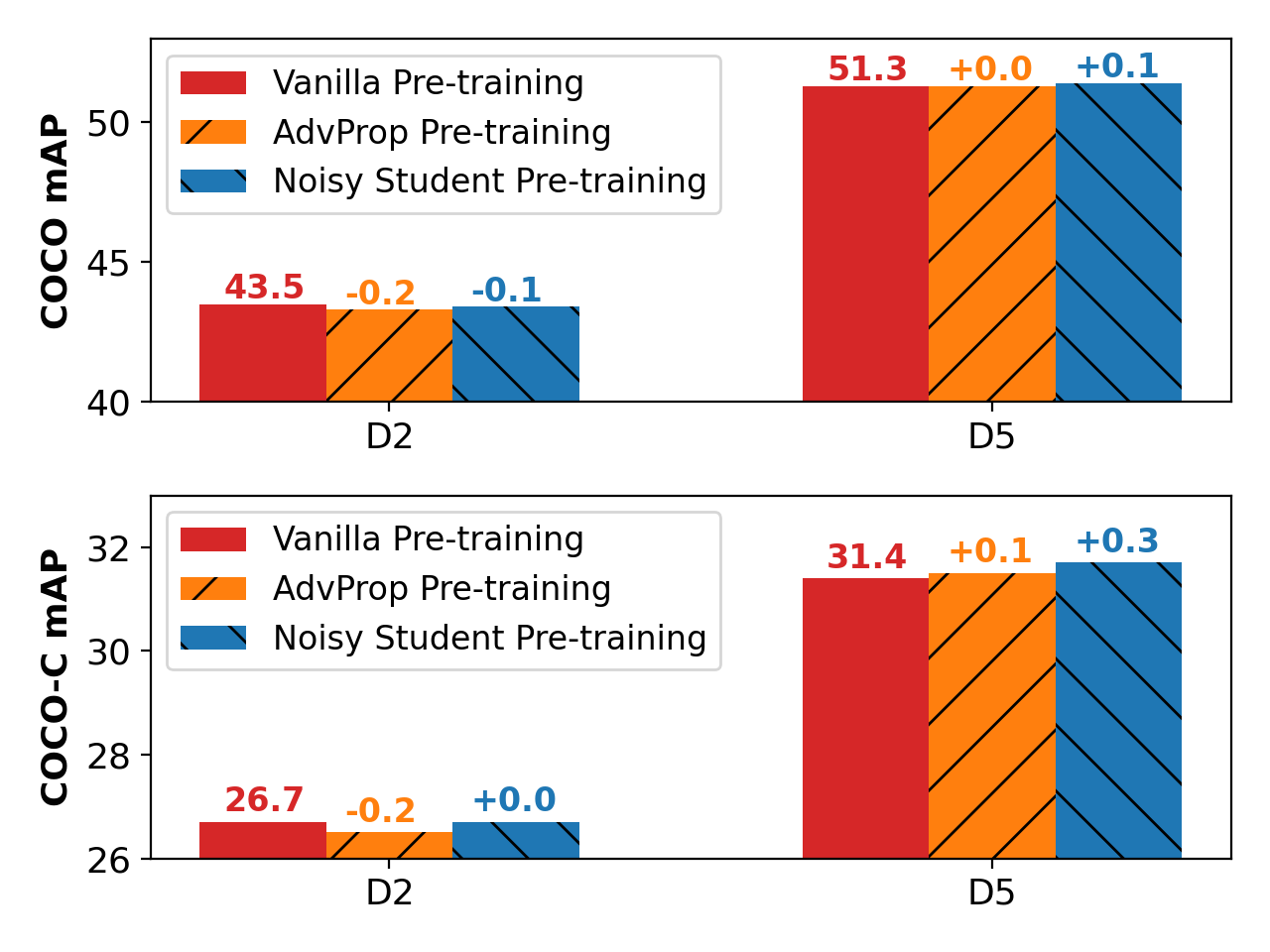}
\caption{Results of EfficientDet-D2 and D5 on COCO and COCO-C with vanilla, AdvProp \cite{xie2020advprop}, and Noisy Student \cite{xie2020self} pre-trained backbones.
Fine-tuning substantially attenuates the effect of these data augmentation methods applied to pre-training.}
\label{fig:pretrain}
\vspace{-12pt}
\end{figure}

We first examine the first choice (i). We follow the common practice to fine-tune pre-trained ImageNet classifiers using the MS COCO object detection dataset~\cite{lin2015coco}. Specifically, we initialize three sets of EfficientDets~\cite{tan2020edet} (each set contains a small-scale detector D2 and a large-scale detector D5) by three ImageNet models, which are respectively pre-trained with no strong augmentation, AdvProp~\cite{xie2020advprop}, and Noisy Student~\cite{xie2020self}. We fine-tune the networks up to 300 epochs to ensure they converge. 

Figure \ref{fig:pretrain} shows the comparison results on the validation sets of MS COCO~\cite{lin2015coco} and COCO-C~\cite{michaelis2020benchmarking}, respectively. The COCO-C dataset provides natural corruption to the COCO images aiming to test a detector's robustness to the real-world adversary. Although the checkpoints produced by AdvProp and Noisy Student surpass their vanilla counterparts (e.g., for B5, AdvProp +0.6\% accuracy on ImageNet and -6.1\% mean corruption error on ImageNet-C~\cite{hendrycks2018benchmarking}), their advantages do not transfer to the object detection.
The detectors built upon them perform similarly and sometimes even worse on COCO than the detector fine-tuned from the vanilla checkpoint. A similar observation holds on COCO-C. Our study is in the same vein as~\cite{he2019rethinking}, but we examine the detectors' robustness in addition to their accuracy.

Since the fine-tuning attenuates the performance gains by strong data augmentation in the upstream classification task, we shift to the second option, 
augmenting during fine-tuning for harvesting data augmentation to improve a detector's accuracy and robustness. It is nontrivial to kill accuracy and robustness with one stone. 
Our first trial is with AutoAugment~\cite{zoph2019learning}, which is searched based on the RetinaNet object detector~\cite{lin2017focal} over COCO. However, it does not generalize well to other detectors (EfficientDets in this work), albeit still on the same COCO dataset (see Tables \ref{tab:main}, \ref{tab:single-class}, \ref{tab:voc} for the results). 

In light of these challenges, we develop a model-dependent augmentation for object detection via adversarial learning, given AdvProp's promising results~\cite{xie2020advprop}. Our method enables a detector to model adversarial images sourced from both object classification and localization, enforcing the detector to learn from its own weaknesses without the need for policy search as in AutoAugment~\cite{zoph2019learning}. 

\section{Approach}
This section describes our main approach to improving the object detectors' robustness and accuracy. We start by reviewing AdvProp for classification~\cite{xie2020advprop} and then tailor it for object detection.

\subsection{AdvProp for Classification Revisited}
Prior works that jointly train with clean and adversarial examples meet with performance degradation on clean images despite the improvement of robustness (e.g., against adversarial attacks)~\cite{madry2018towards, kurakin2017adversarial, zhang2019trades}.
Most recently, Xie \etal.\ manage to leverage the adversarial examples to improve image classifiers' accuracy, besides robustness,  by a new training paradigm named Adversarial Propagation (AdvProp)~\cite{xie2020advprop}.
Intuitively, they argue that the clean and adversarial images are drawn from distinct distributions, making it suboptimal to share the same statistical estimation in batch normalization (batchnorm) layers.
To disentangle the two underlying distributions, AdvProp introduces auxiliary running mean and running variance for the adversarial images, leaving the main batchnorm layers to serve solely for the clean images.
In particular, in every epoch, they first generate adversarial images based on the auxiliary batchnorm.
Then they forward the clean and adversarial mini-batches into the network, each with their exclusive batchnorm, followed by standard backpropagation to optimize the total loss.

Under such a training scheme, the separate batchnorms take care of the distribution shift between clean and adversarial examples, jointly giving rise to higher performance than learning only with the clean images. 
Another benefit is robustness. The models trained by AdvProp is more robust against image distortions, probably because they are immune to the adversarially distorted examples by training.
Inspired by their success in image classification, we propose Det-AdvProp to build accurate and robust detectors as object detection plays a crucial role in many real-world applications such as autonomous driving.

\subsection{Det-AdvProp}
\label{sec:det-advprop}
One-stage object detectors take an image $\textbf{x}$ as input and predict a set of objects $\{(\hat{\textbf{y}},\hat{\textbf{b}})\}$, each includes a vector $\hat{\textbf{y}}$ representing the probabilities over all possible classes and 4-dimensional bounding box coordinates $\hat{\textbf{b}}$.
During training, the objective function is usually formulated as follows:
\begin{align}
\min_\theta\ \mathbb{E}_{\textbf{x}\sim\mathcal{D}; y,\textbf{b}\sim\mathcal{B}(\textbf{x})}&\ L_{det}(\textbf{x}, y,\textbf{b}; \theta), \\
L_{det}(\textbf{x}, y,\textbf{b}; \theta) =& L_{cls}(\textbf{x}, y; \theta) + w \cdot L_{loc}(\textbf{x}, \textbf{b}; \theta), \label{eq:loss}
\end{align}
where $\mathcal{D}$ is the set of training images, $\mathcal{B}(\textbf{x})$ collects all the class $y$ and bounding box $\textbf{b}$ labels of the objects in image $\textbf{x}$, $\theta$ is the model parameter, and $w$ is the weight to balance the classification loss $L_{cls}$ (e.g., focal loss \cite{lin2017focal}) and the localization loss $L_{loc}$ (e.g., Huber loss).

Equation~(\ref{eq:loss}) essentially defines a multi-task learning problem, so possible attacks on object detectors can emerge from two heterogeneous domains for classification and localization, respectively. In some sense, it is easier to attack object detectors  than image classifiers. Many existing methods have successfully fooled popular object detectors by attacking an individual loss in equation~(\ref{eq:loss})~\cite{xie2017adversarial, chen2019shapeshifter, kevin2018phyattack}.

It is a double-edged sword to have the adversarial examples of heterogeneous domains for object detectors. On the one hand, they may be viewed as versatile augmentations to the training data, potentially improving the detectors' accuracy on clean test data.
On the other hand, the detectors need to be enhanced against both domains of adversarial examples to achieve robustness. 

\vspace{-10pt}
\paragraph{AdvProp over $L_{det}$.}
Due to the multi-task nature of the detectors, a straightforward application of AdvProp to object detection, i.e., by attacking the total loss $L_{det}$ and dedicating two batchnorms, does not work well (see Table~\ref{tab:taskloss}). Its performance is about the same or worse than attacking the single-task classification loss $L_{cls}$. We can further understand this observation following~\cite{zhang2019towards}, which finds that the two gradients, $\nabla_\textbf{x}L_{cls}$ and $\nabla_\textbf{x}L_{loc}$ which are generated when we attack the overall detection loss $L_{det}$, have distinct value ranges and inconsistent directions. They reduce the augmentation effect of the resultant adversarial examples and even mutually cancel out.

\vspace{-10pt}
\paragraph{AdvProp over $L_{cls}$ and $L_{loc}$.}
Another plausible approach is to model the clean images and the attacks sourced from classification and localization as three distinct domains. To disentangle them, we use three batchnorms. One reserved for clean images, another accounting for the adversarial examples generated by attacking the classification loss, and the last for the localization-sourced adversarial examples.
However, this method hurts the detectors' performance on clean images though it yields the highest robustness on corrupted images (see Table \ref{tab:taskloss}). We conjecture the two auxiliary batchnorms are overly strong regularization during training, leading to the detector's under-fitting to the clean training set.

\begin{algorithm}[!t]
\caption{Det-AdvProp}
\KwInput{Object detection dataset $\mathcal{D}$}
\KwOutput{Learned network parameter $\theta$}
\begin{algorithmic}[1]
\FOR{each training epoch}
\STATE Sample a random batch $\{\textbf{x}^i,\{y^i,\textbf{b}^i\}\}\sim \mathcal{D}$
\STATE Generate $\hat{\textbf{x}}^i_{cls}$ based on $L_{cls}(\textbf{x}^i,y^i)$ using \\ auxiliary batchnorm
\STATE Generate $\hat{\textbf{x}}^i_{loc}$ based on $L_{loc}(\textbf{x}^i,\textbf{b}^i)$ using \\ auxiliary batchnorm
\STATE Select $\hat{\textbf{x}}^i$ based on Eq \eqref{eq:mma}
\STATE Compute $L_{det}(\textbf{x}^i,\{y^i,\textbf{b}^i\})$ with main batchnorm
\STATE Compute $L_{det}(\hat{\textbf{x}}^i,\{y^i,\textbf{b}^i\})$ with auxiliary batchnorm
\STATE Perform a step of gradient descent w.r.t. $\theta$ \\ 
$\min\ L_{det}(\textbf{x}^i,y^i,\textbf{b}^i)+L_{det}(\hat{\textbf{x}}^i,y^i,\textbf{b}^i)$
\ENDFOR
\end{algorithmic}
\label{alg:schema}
\end{algorithm}

\begin{table*}[!htb]
\centering
\resizebox{.68\textwidth}{!}{
\begin{tabular}{l|c|c|c|c|c|c}
\hline
Model & mAP & AP50 & AP75 & APl & APm & APs \\
\hline\hline
\textbf{EfficientDet-D0} & 34.3 & 52.4 & 36.6 & 53.8 & 40.0 & 13.1 \\
+ AutoAugment & 34.4 (+0.1) & 52.8 (+0.4) & 36.7 (+0.1) & 53.1 (-0.7) & 40.2 (+0.2) & \textbf{13.9 (+0.8)} \\
+ Det-AdvProp (ours) & \textbf{34.7 (+0.4)} & \textbf{52.9 (+0.5)} & \textbf{37.2 (+0.6)} & \textbf{54.1 (+0.3)} & \textbf{40.6 (+0.6)} & \textbf{13.9 (+0.8)} \\ \hline
\textbf{EfficientDet-D1} & 40.2 & 58.8 & 42.8 & 57.7 & 45.9 & \textbf{21.2} \\
+ AutoAugment & 40.1 (-0.1) & \textbf{59.2 (+0.4)} & 43.2 (+0.4) & 57.9 (+0.2) & 45.7 (-0.2) & 19.9 (-1.2) \\
+ Det-AdvProp (ours) & \textbf{40.5 (+0.3)} & \textbf{59.2 (+0.4)} & \textbf{43.3 (+0.5)} & \textbf{58.8 (+1.1)} & \textbf{46.2 (+0.3)} & 20.6 (-0.6) \\ \hline
\textbf{EfficientDet-D2} & 43.5 & 62.5 & 47.1 & 60.9 & 48.6 & 23.7 \\
+ AutoAugment & 43.5 (+0.0) & \textbf{62.8 (+0.3)} & 46.6 (-0.5) & 59.8 (-1.1) & 48.7 (+0.1) & 23.9 (+0.2) \\
+ Det-AdvProp (ours) & \textbf{43.8 (+0.3)} & 62.6 (+0.1) & \textbf{47.3 (+0.2)} & \textbf{61.0 (+0.1)} & \textbf{49.6 (+1.0)} & \textbf{25.6 (+1.9)} \\ \hline
\textbf{EfficientDet-D3} & 46.8 & 65.3 & 50.6 & 62.8 & 51.6 & 29.8 \\
+ AutoAugment & 47.0 (+0.2) & 66.0 (+0.7) & 50.8 (+0.2) & 63.0 (+0.2) & 51.7 (+0.1) & 29.8 (+0.0) \\
+ Det-AdvProp (ours) & \textbf{47.6 (+0.8)} & \textbf{66.3 (+1.0)} & \textbf{51.4 (+0.8)} & \textbf{64.0 (+1.2)} & \textbf{52.2 (+0.6)} & \textbf{30.2 (+0.4)} \\ \hline
\textbf{EfficientDet-D4} & 49.3 & 68.2 & 53.3 & 63.7 & 53.6 & \textbf{33.0} \\
+ AutoAugment & 49.5 (+0.2) & \textbf{68.7 (+0.5)} & 53.7 (+0.4) & 64.9 (+1.2) & 54.0 (+0.4) & 31.9 (-1.1) \\
+ Det-AdvProp (ours) & \textbf{49.8 (+0.5)} & 68.6 (+0.4) & \textbf{54.2 (+0.9)} & \textbf{65.2 (+1.5)} & \textbf{54.2 (+0.6)} & 32.4 (-0.6) \\ \hline
\textbf{EfficientDet-D5} & 51.3 & 70.1 & 55.8 & 65.1 & 55.1 & 35.9 \\ 
+ AutoAugment & 51.5 (+0.2) & 70.4 (+0.3) & 56.0 (+0.2) & 65.2 (+0.1) & 56.1 (+1.0) & 35.4 (-0.5) \\
+ Det-AdvProp (ours) & \textbf{51.8 (+0.5)} & \textbf{70.7 (+0.6)} & \textbf{56.3 (+0.5)} & \textbf{66.1 (+1.0)} & \textbf{56.2 (+1.1)} & \textbf{36.2 (+0.3)} \\ \hline
\end{tabular}}
\smallskip
\caption{Comparison of vanilla training, AutoAugment~\cite{zoph2019learning}, and Det-AdvProp on MS COCO~\cite{lin2015coco}. Our proposed Det-AdvProp consistently outperforms vanilla training  for different detectors, and it performs better than AutoAugment on all object sizes.}
\label{tab:main}
\vspace{-10pt}
\end{table*}

\vspace{-10pt}
\paragraph{Det-AdvProp.}
Learning from the two lessons above, we develop the following training scheme named Det-AdvProp. Similar to the AdvProp over $L_{det}$,  we still use one, not two, auxiliary batchnorm to take account of the adversarial examples. Similar to the AdvProp over  $L_{cls}$ and $L_{loc}$, we generate two adversarial examples for each input by attacking the two losses separately. However, we keep only one of them to avoid the potential conflict between the adversarial gradients of the classification loss and the localization loss by a max-max rule first proposed in~\cite{zhang2019towards}.

More concretely, we use the FGSM algorithm~\cite{goodfellow2015fgsm} with non-targeted attack to explain Det-AdvProp without loss of generality. Given an input image $\textbf{x}$ and a bounding box $\textbf{b}$ over an object of class $y$, FGSM produces the adversarial examples by a one-step projected gradient descent:
\begin{align}
\hat{\textbf{x}}_{cls} &= \mathcal{P}(\textbf{x}+\epsilon \cdot \text{sign}(\nabla_{\textbf{x}}L_{cls}(\textbf{x}, y; \theta))), \\
\hat{\textbf{x}}_{loc} &= \mathcal{P}(\textbf{x}+\epsilon \cdot \text{sign}(\nabla_{\textbf{x}}L_{loc}(\textbf{x}, \textbf{b}; \theta))), \\
\hat{\textbf{x}} &= \argmax_{\textbf{x}\in \{\hat{\textbf{x}}_{cls}, \hat{\textbf{x}}_{loc}\}}L_{det}(\textbf{x},y,\textbf{b};\theta),
\label{eq:mma}
\end{align}
where $\epsilon$ is the attack strength and $\mathcal{P}$ denotes the projection onto the norm ball $\{\hat{\textbf{x}}\mid\|\hat{\textbf{x}}-\textbf{x}\|_\infty \leq \epsilon\}$.
We first generate two adversarial examples by maximizing the single task loss and then choose the one that maximizes the total loss of the detection task. Equation~(\ref{eq:mma}) is the max-max rule to keep the adversarial example (out of two) that maximizes the total detection loss $L_{det}$. The inner ``max'' in the max-max rule refers to the adversarial examples that maximize $L_{cls}$ and $L_{loc}$, respectively, and the outer ``max'' indicates we take the one that maximizes the total detection loss $L_{det}$.
The overall training objective is given below:
\begin{align}
    \min_\theta\ & \mathbb{E}_{\textbf{x}\sim\mathcal{D}; y,\textbf{b}\sim\mathcal{B}(\textbf{x})} L_{det}(\textbf{x},y,\textbf{b};\theta) + L_{det}(\hat{\textbf{x}},y,\textbf{b};\theta).
\end{align}


Intuitively, this max-max scheme lets the stronger one survive between the two  adversarial examples $\hat{\textbf{x}}_{cls}$ and $\hat{\textbf{x}}_{loc}$.  
By separately attacking the classification and localization branches, we avoid the gradient misalignment problem.
Since we only produce one adversarial example per clean image, coupled with one auxiliary batchnorm, the training does not suffer from the excessive regularization problem mentioned earlier.

Algorithm~\ref{alg:schema} describes Det-AdvProp in detail.
Note that we can safely discard the auxiliary batchnorm after training and use the main batchnorm for inference. The object detectors trained by Det-AdvProp have the same parameters and latency as those obtained by vanilla training.

\begin{figure*}[!htb]
\centering
\resizebox{.96\textwidth}{!}{
\subfloat[]{\label{fig:combine}\includegraphics[width=0.5\linewidth]{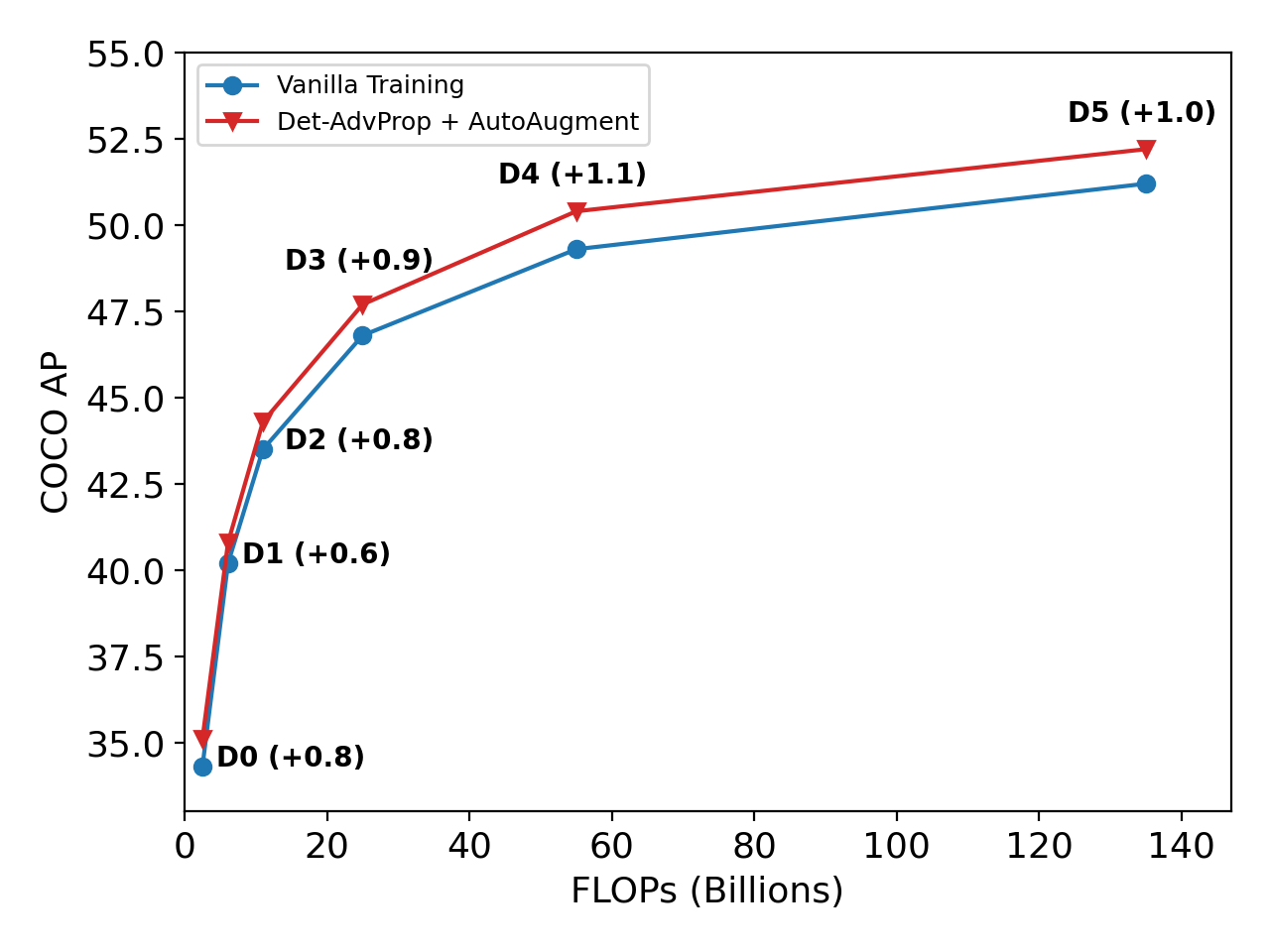}}
\subfloat[]{\label{fig:coco-c-aa}\includegraphics[width=0.5\linewidth]{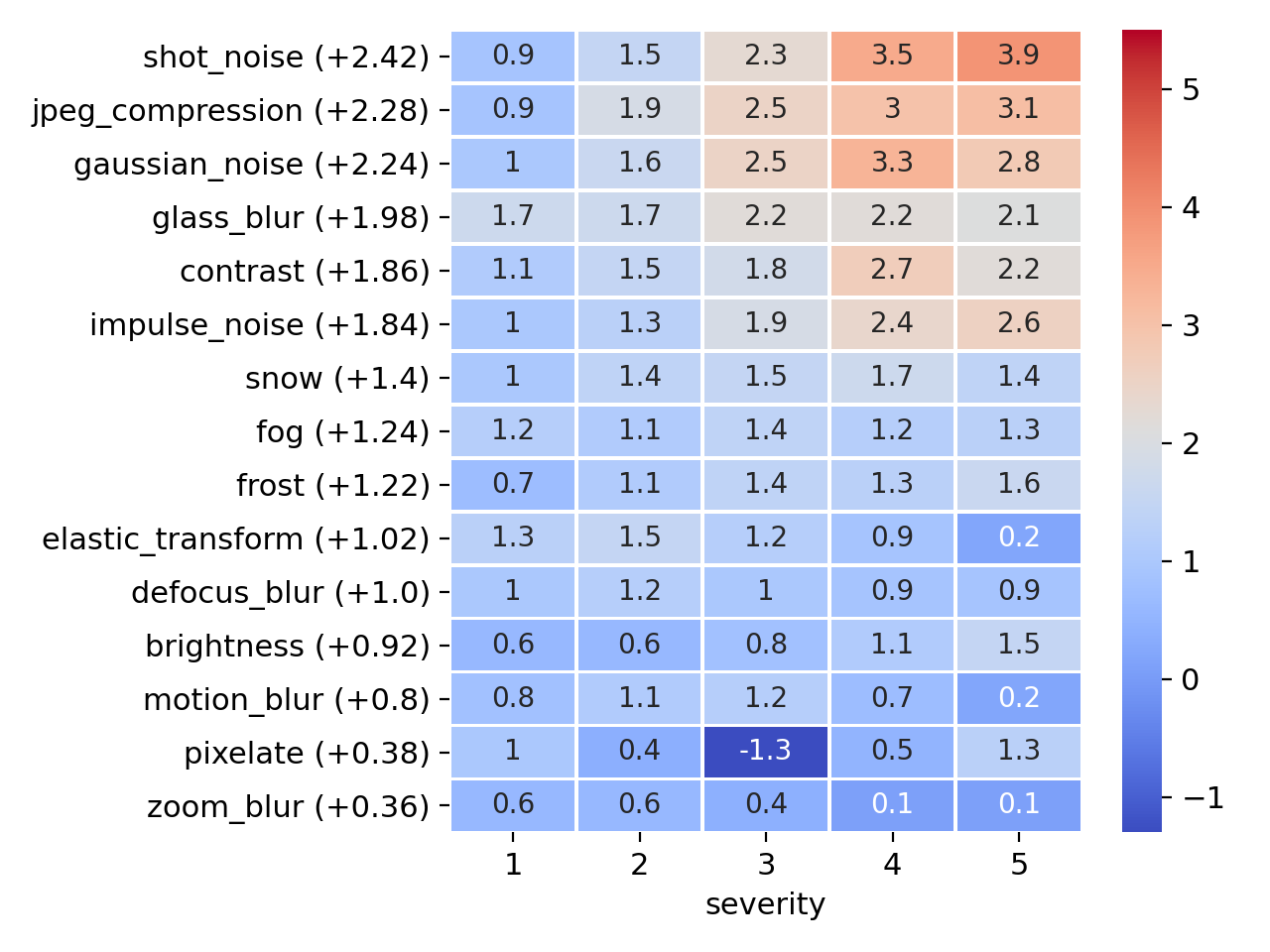}}
\subfloat[]{\label{fig:coco-c-ap}\includegraphics[width=0.5\linewidth]{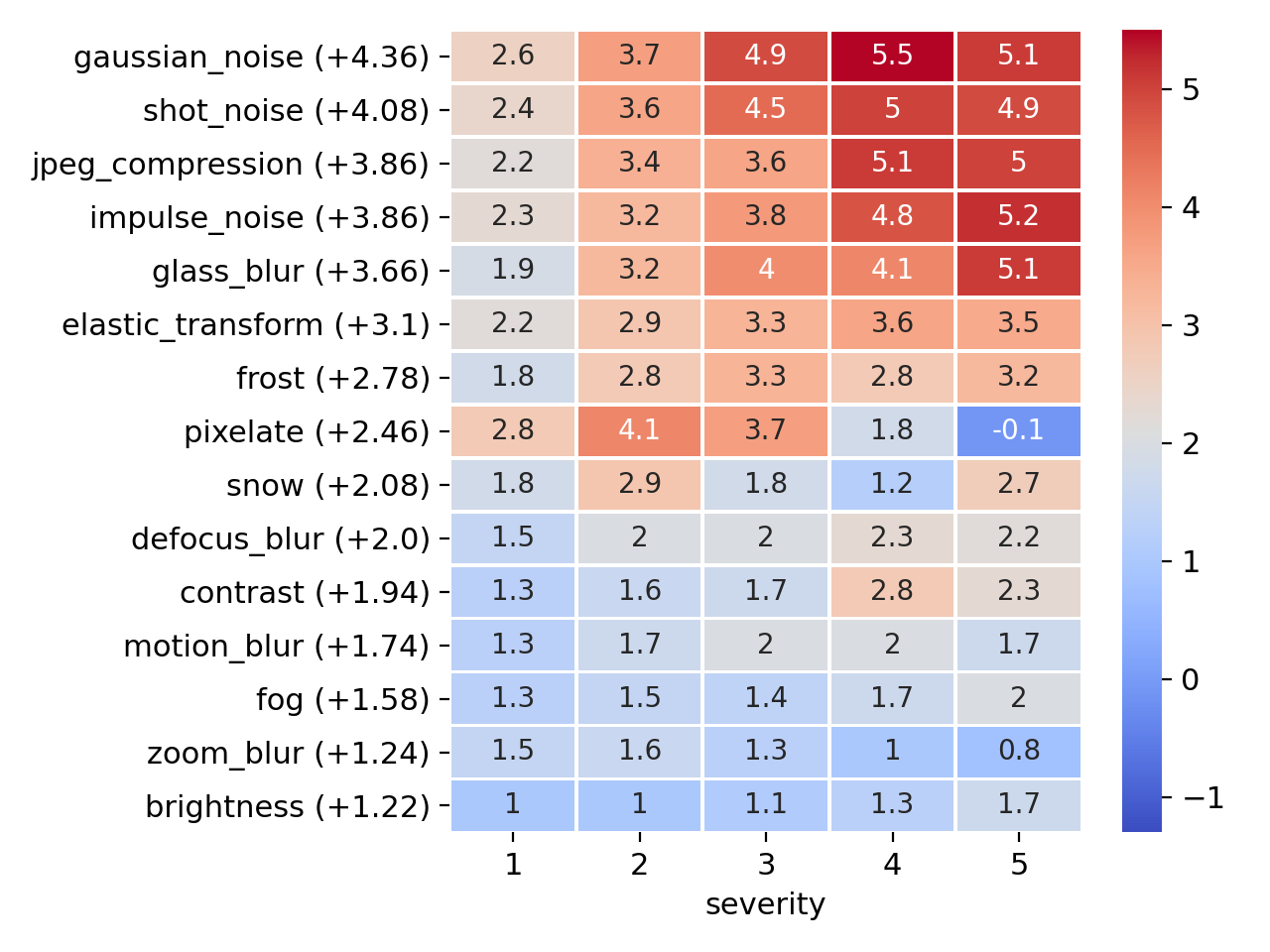}}
\vspace{-10pt}
}
\caption{\textbf{Left:} Combining Det-AdvProp with AutoAugment \cite{zoph2019learning} can produce slightly better results than using Det-AdvProp alone.
EfficientDet-D4 achieves over 50 mAP with 21M parameters, which is 10x less than AmoebaNet+NAS with NAS-FPN \cite{ghiasi2019nasfpn} and AutoAugment \cite{zoph2019learning}.
\textbf{Middle:} Performance gains of AutoAugment over vanilla training on COCO-C \cite{michaelis2020benchmarking}.
\textbf{Right:} Performance gains of Det-AdvProp over vanilla training on COCO-C. The largest improvement is observed when the images are distorted by random noise and with strong corruption strength.
(best viewed in color).}
\vspace{-12pt}
\end{figure*}

\section{Experiments}

\subsection{Setups}
We select EfficientDet~\cite{tan2020edet} of various scales as the default object detectors, including the lightweight EfficientDet-D0 model with fewer than 4M parameters and the large-scale EfficientDet-D5 detector that achieves over 50 mAP on COCO~\cite{lin2015coco}.
We train the detectors using the COCO 2017 object detection dataset~\cite{lin2015coco} for 300 epochs and evaluate them on COCO's validation set to obtain the so called clean accuracy. We also test the detectors' robustness to natural corruptions on the COCO-C dataset~\cite{michaelis2020benchmarking}, including 15 types of corruption each with 5 severity levels. Finally, we apply the detectors to the PASCAL VOC 2012 dataset~\cite{pascal-voc-2012} to evaluate robustness under domain shift. Please refer to Appendix~\ref{app:settings} for details and training complexity.

\subsection{Det-AdvProp Improves Accuracy}
\label{sec:acc}
\paragraph{Results on COCO.}
We train the EfficientDet detectors with Det-AdvProp and compare them with the models out of vanilla training and the AutoAugment searched for object detection~\cite{zoph2019learning} in Table~\ref{tab:main}.
As different scales of detectors favor different attack manners (targeted vs.\ non-targeted), which we ablate next, we first report the best clean mAP here. See Appendix~\ref{app:settings} for the corresponding attack manners. 
Compared with the vanilla training baseline, our method can consistently increase the mAP score over all the detectors of various scales. The improvement mainly comes from large and medium-sized objects.
The performance gain is especially notable for large-scale detectors with high capacities. Det-AdvProp improves D0--D2 by at most 0.4 mAP, but boosts D3---D5 by at least 0.5 mAP. 
On the contrary, AutoAugment cannot transfer well to the EfficientDets.
It even incurs performance drop for some models (e.g., -0.1 mAP on D1).
Det-AdvProp outperforms AutoAugment not only for the overall results but also on all the sizes of objects.


To push the limit of our method, we further combine the two strategies by employing AutoAugment to fine-tune the detectors obtained by Det-AdvProp. According to Figure \ref{fig:combine}, the combined augmentation leads to even more accurate detectors.
Similar to the previous observations, the performance gain is proportional to the model capacity. 
Detectors larger than D3 all increase by at least 1.0 mAP.
With the combined augmentation strategy, D4 can achieve 50.4 mAP with just 21M parameters.
As a comparison, AmoebaNet accompanied by NAS-FPN \cite{ghiasi2019nasfpn} and AutoAugment \cite{zoph2019learning} achieves 50.7 mAP but has 209M parameters, which is 10x larger than D4.
The analysis in Appendix~\ref{app:analysis} further shows that we can achieve performance gain on most classes and reduce various types of detection errors.

\begin{table}[!htb]
\centering
\resizebox{.44\textwidth}{!}{
\begin{tabular}{c|c|c|c|c|c|c}
\hline
Model & D0 & D1 & D2 & D3 & D4 & D5 \\ \hline \hline
Vanilla & 33.4 & 36.6 & 38.2 & 41.7 & 44.6 & 46.5 \\ \hline
Det-AdvProp & \tabincell{c}{\textbf{33.9} \\ \textbf{(+0.5)}} & \tabincell{c}{\textbf{37.9} \\ \textbf{(+1.3)}} & \tabincell{c}{\textbf{40.0} \\ \textbf{(+1.8)}} & \tabincell{c}{\textbf{43.8} \\ \textbf{(+2.1)}} & \tabincell{c}{\textbf{46.2} \\ \textbf{(+1.6)}} & \tabincell{c}{\textbf{48.2} \\ \textbf{(+1.7)}} \\ \hline
\end{tabular}}
\smallskip
\caption{Performance comparison between Det-AdvProp and the vanilla baseline without ImageNet pre-training. 
Det-AdvProp gives rise to performance gains for all detectors.}
\label{tab:no-pretrain}
\vspace{-10pt}
\end{table}

\vspace{-10pt}
\paragraph{Without Pre-training.} 
In some real-world settings, the ImageNet pre-trained backbone is not always accessible due to various reasons.
Previous work also shows that training object detectors from scratch can match the performance with pre-training given sufficiently many training iterations~\cite{he2019rethinking}.
Hence, we also report the improvement of our method over the vanilla training baseline when EfficientDets are initialized from scratch.
The other training settings are exactly the same as before.
As illustrated in Table~\ref{tab:no-pretrain}, the proposed Det-AdvProp's effectiveness is magnified in this scenario. 
The augmented models can be over 2.0 mAP better than those trained via the vanilla baseline.

\vspace{-10pt}
\paragraph{Single-class object detection.}
In certain applications, there is only one object class of interest. They desire a detector to localize that object out of the background, such as face detection, pedestrian detection, etc.
To simulate this scenario, we choose three classes spanning different object sizes and numbers of training instances from the COCO dataset to test Det-AdvProp under the single-class object detection setting. 
We follow exactly the same experiment settings as before and always use the attack strength of $\epsilon=1$.
Table~\ref{tab:single-class} shows the results of EfficientDet-D3, where AutoAugment again degrades the mAP for every class.
Its policy is searched based on 80 classes and fails to adapt well to the single-class object detection.
In contrast, Det-AdvProp enhances the detector by automatically learning from its own weakness and thus achieves consistent improvement.
The performance gain tends to be larger when there are fewer training images.

\begin{table}[!htb]
\centering
\resizebox{.4\textwidth}{!}{
\begin{tabular}{c|c|c||c|c|c}
\hline
Class & \tabincell{c}{Object \\ Size} & \tabincell{c}{\# Images} & Vanilla & \tabincell{c}{Auto-\\ Augment} & \tabincell{c}{Det-AdvProp \\ (ours)} \\ \hline \hline
\textbf{Donut} & Small & 1,585 & 25.4 & \tabincell{c}{23.9 \\ (-1.5)} & \tabincell{c}{\textbf{28.7} \\ \textbf{(+3.3)}} \\ \hline
\textbf{Person} & Medium & 66,808 & 58.2 & \tabincell{c}{58.0 \\ (-0.2)} & \tabincell{c}{\textbf{58.5} \\ \textbf{(+0.3)}} \\ \hline
\textbf{Truck} & Large & 6,377 & 28.1 & \tabincell{c}{25.5 \\ (-2.6)} & \tabincell{c}{\textbf{28.7} \\
\textbf{(+0.6)}} \\ \hline
\end{tabular}}
\smallskip
\caption{mAP of vanilla, AutoAugment~\cite{zoph2019learning}, and Det-AdvProp applied to EfficientDet-D3 under the single-class object detection setting.
The AutoAugement strategy is searched based on 80 classes and fails to adapt to this setting.
Det-AdvProp maintains its effectiveness in part due to its model-dependent nature.}
\label{tab:single-class}
\vspace{-10pt}
\end{table}

\subsection{Det-AdvProp Improves Robustness}
\label{sec:robust}
\paragraph{Results on distorted COCO-C.}
We further evaluate the robustness of the detectors trained by Det-AdvProp under various corruptions on COCO-C \cite{michaelis2020benchmarking}.
Detecting objects from COCO-C is much more difficult than that from clean images.
The popular Faster-RCNN \cite{ren2015faster} model with ResNet50 \cite{he2016resnet} backbone achieves 36.3 mAP on clean COCO images, while its mAP reduces to only 18.2 on COCO-C.
As shown in Table~\ref{tab:coco-c}, Det-AdvProp can achieve more significant improvement on distorted images than on the clean images over the baseline. 
For instance, Det-AdvProp can improve EfficientDet-D4 by 2.7 mAP on COCO-C, which is over 5x of the improvement on clean COCO.
Compared with AutoAugment, we can consistently double its robustness improvement on various scales of EfficientDets.
To further break down the improvement into different corruptions and severities, we visualize the performance gain achieved by AutoAugment~\cite{zoph2019learning} and the proposed Det-AdvProp in Figures~\ref{fig:coco-c-aa} and \ref{fig:coco-c-ap}.
Models trained by Det-AdvProp outperform the vanilla models on all 15 corruptions, and we observe the largest improvement when the images are distorted by random noise (e.g., +4.36 mAP under Gaussian noise and +4.08 mAP under shot noise).
Another interesting finding is that the performance gains tend to become larger when the corruption strengths are stronger.
Although AutoAugment can also help the models generalize to COCO-C, it is less effective than ours against every type of corruptions.
The results of combined Det-AdvProp and AutoAugment is shown in Appendix \ref{app:coco-c}, where we achieve the largest improvement of +3.8 mAP.

\begin{table}[!htb]
\centering
\resizebox{.4\textwidth}{!}{
\begin{tabular}{l|c|c|c}
\hline
Model & Vanilla & AutoAugment & Det-AdvProp (ours) \\
\hline\hline
\textbf{EfficientDet-D0} & 21.4 & 21.8 (+0.4) & \textbf{22.2 (+0.8)} \\
\textbf{EfficientDet-D1} & 24.4 & 25.1 (+0.7) & \textbf{25.6 (+1.2)} \\ 
\textbf{EfficientDet-D2} & 26.7 & 27.1 (+0.4) & \textbf{27.6 (+0.9)} \\ 
\textbf{EfficientDet-D3} & 28.8 & 29.6 (+0.8) & \textbf{30.8 (+2.0)} \\ 
\textbf{EfficientDet-D4} & 30.1 & 31.5 (+1.4) & \textbf{32.8 (+2.7)} \\
\textbf{EfficientDet-D5} & 31.4 & 32.6 (+1.2) & \textbf{33.7 (+2.3)} \\ \hline
\end{tabular}}
\smallskip
\caption{Comparison of augmentation strategies and vanilla training on COCO-C~\cite{michaelis2020benchmarking}. 
Det-AdvProp can double the improvement achieved by AutoAugment~\cite{zoph2019learning}.}
\label{tab:coco-c}
\vspace{-10pt}
\end{table}

\vspace{-10pt}

\paragraph{Results on cross-dataset generalization.}
Another manifest of the model robustness is whether it can retain strong performance against domain shift.
PASCAL VOC 2012 \cite{pascal-voc-2012} only contains 20 classes, which are much smaller than the 80 labeled classes in COCO. 
The underlying distributions of the two datasets are also different in the image content or the bounding box sizes and locations.
We use the trained detectors to run inference directly on the VOC dataset to test their transferibility. We maintain the COCO evaluation metrics in this experiment.
According to Table \ref{tab:voc}, the Det-AdvProp trained detectors always outperform those by vanilla and AutoAugment training under every model scale and every evaluation metric.
The models obtained by AutoAugment even substantially underperform the vanilla models.
For instance, AutoAugment lowers D2's mAP by 0.6 compared to the vanilla training baseline.
We also show the results of Det-AdvProp + AutoAugment in Appendix \ref{app:voc}, where the largest improvement is +1.3 mAP on EfficientDet-D5.

\begin{table}[!htb]
\vspace{-4pt}
\centering
\resizebox{.38\textwidth}{!}{
\begin{tabular}{l|c|c|c}
\hline
Model & mAP & AP50 & AP75 \\
\hline\hline
\textbf{EfficientDet-D0} & 55.6 & 77.6 & 61.4 \\
+ AutoAugment & 55.7 (+0.1) & 77.7 (+0.1) & 61.8 (+0.4) \\
+ Det-AdvProp (ours) & \textbf{55.9 (+0.3)} & \textbf{77.9 (+0.3)} & \textbf{62.0 (+0.6)} \\ \hline
\textbf{EfficientDet-D1} & 60.8 & 82.0 & 66.7 \\
+ AutoAugment & 61.0 (+0.2) & 82.2 (+0.2) & 67.2 (+0.5) \\
+ Det-AdvProp (ours) & \textbf{61.2 (+0.4)} & \textbf{82.3 (+0.3)} & \textbf{67.4 (+0.7)} \\ \hline
\textbf{EfficientDet-D2} & 63.3 & 83.6 & 69.3 \\
+ AutoAugment & 62.7 (-0.6) & 83.3 (-0.3) & 69.2 (-0.1) \\
+ Det-AdvProp (ours) & \textbf{63.5 (+0.2)} & \textbf{83.8 (+0.2)} & \textbf{69.7 (+0.4)} \\ \hline
\textbf{EfficientDet-D3} & 65.7 & 85.3 & 71.8 \\
+ AutoAugment & 65.2 (-0.5) & 85.1 (-0.2) & 71.3 (-0.5) \\
+ Det-AdvProp (ours) & \textbf{66.2 (+0.5)} & \textbf{85.9 (+0.6)} & \textbf{72.5 (+0.7)} \\ \hline
\textbf{EfficientDet-D4} & 67.0 & 86.0 & 73.0 \\
+ AutoAugment & 67.0 (+0.0) & 86.3 (+0.3) & 73.5 (+0.5) \\
+ Det-AdvProp (ours) & \textbf{67.5 (+0.5)} & \textbf{86.6 (+0.6)} & \textbf{74.0 (+1.0)} \\ \hline
\textbf{EfficientDet-D5} & 67.4 & 86.9 & 73.8 \\
+ AutoAugment & 67.6 (+0.2) & 87.2 (+0.3) & 74.2 (+0.4) \\
+ Det-AdvProp (ours) & \textbf{68.2 (+0.8)} & \textbf{87.6 (+0.7)} & \textbf{74.7 (+0.9)} \\ \hline
\end{tabular}}
\smallskip
\caption{Results on PASCAL VOC 2012.
The proposed Det-AdvProp gives the highest score on every model and metric.
It largely outperforms AutoAugment~\cite{zoph2019learning} when facing domain shift.
}
\vspace{-12pt}
\label{tab:voc}
\end{table}

\subsection{Ablation Study}
\label{sec:ablation}
\paragraph{Det-AdvProp with targeted and non-targeted attacks.}
Targeted attack aims to fool a model to recognize an image incorrectly as a specified target label, while non-targeted attack is conducted by maximizing the training loss on the true label.
We first carry out targeted attack with Det-AdvProp.
As previous works report performance improvement if the object and background are treated differently \cite{zoph2019learning}, we consider the following two ways to generate random targets:
(i) Randomly generate target labels for all the predefined anchors in the detector; (ii) Only perturb the ground-truth label for the anchors that cover objects, omitting those background anchors.
We find that (ii) performs almost the same as vanilla training even when the attack strength is very large ($\epsilon=5$).
It is probably because the adversarial images which fool models to misidentify the background as objects contain valuable features, and the number of object-covering anchors is considerably smaller than the background anchors.
Hence, for the targeted attack mentioned below, we refer to generating adversarial labels for all anchors if not specified otherwise.
For the non-targeted attack, we simply maximize the training loss for both classification and localization branches and use the same attack strength ($\epsilon=1$) as the targeted attack for a fair comparison.

The results are shown in Table \ref{tab:tg-ntg}.
On clean COCO images, the adversarial examples obtained by targeted attack consistently improve the models' mAP, but non-targeted attack can hurt the performance of lightweight detectors (D0-D2), implying possibly too strong regularization.
In contrast, when it comes to larger models (D3-D5), Det-AdvProp works better with the relatively stronger non-targeted attack.
On the COCO-C dataset, both attack methods can improve the detectors' robustness, but the improvement achieved by non-targeted attack is much larger, and the gap is wider for the models of higher capacities.
The relative robustness achieved by non-targeted attack is also the highest for all the detectors.
Intuitively, Det-AdvProp with non-targeted attack learns from the worse-case adversarial examples than the targeted attack within the $\epsilon$ norm ball and therefore provides stronger regularization and robustness to the detectors.

\begin{table}[!htb]
\vspace{-2pt}
\centering
\resizebox{.38\textwidth}{!}{
\begin{tabular}{l|c|c|c}
\hline
Model & \tabincell{c}{COCO \\ mAP} & \tabincell{c}{COCO-C \\ mAP} & \tabincell{c}{Relative \\ rPC (\%)} \\
\hline\hline
\textbf{EfficientDet-D0} & 34.3 & 21.4 & 62.4 \\
+ Det-AdvProp (TG) & \textbf{34.7 (+0.4)} & \textbf{22.2 (+0.8)} & 64.0 (+1.6) \\ 
+ Det-AdvProp (NTG) & 34.0 (-0.3) & 22.1 (+0.7) & \textbf{65.0 (+2.6)} \\ \hline
\textbf{EfficientDet-D1} & 40.2 & 24.4 & 60.7 \\ 
+ Det-AdvProp (TG) & \textbf{40.5 (+0.3)} & 25.6 (+1.2) & 63.2 (+2.5) \\ 
+ Det-AdvProp (NTG) & 40.1 (-0.1) & \textbf{26.1 (+1.7)} & \textbf{65.1 (+4.4)} \\ \hline
\textbf{EfficientDet-D2} & 43.5 & 26.7 & 61.4 \\ 
+ Det-AdvProp (TG) & \textbf{43.8 (+0.3)} & 27.6 (+0.9) & 63.0 (+1.6) \\ 
+ Det-AdvProp (NTG) & 43.4 (-0.1) & \textbf{28.0 (+1.3)} & \textbf{64.5 (+3.1)} \\ \hline
\textbf{EfficientDet-D3} & 46.8 & 28.8 & 61.5 \\ 
+ Det-AdvProp (TG) & 47.2 (+0.4) & 30.1 (+1.3) & 63.8 (+2.3) \\
+ Det-AdvProp (NTG) & \textbf{47.6 (+0.8)} & \textbf{30.8 (+2.0)} & \textbf{64.7 (+3.2)} \\ \hline
\textbf{EfficientDet-D4} & 49.3 & 30.1 & 61.1 \\
+ Det-AdvProp (TG) & 49.6 (+0.3) & 31.8 (+1.7) & 64.1 (+3.0) \\
+ Det-AdvProp (NTG) & \textbf{49.8 (+0.5)} & \textbf{32.8 (+2.7)} & \textbf{65.9 (+4.8)} \\ \hline
\textbf{EfficientDet-D5} & 51.3 & 31.4 & 61.2 \\
+ Det-AdvProp (TG) & 51.5 (+0.2) & 32.4 (+1.0) & 62.9 (+1.7) \\
+ Det-AdvProp (NTG) & \textbf{51.8 (+0.5)} & \textbf{33.7 (+2.3)} & \textbf{65.1 (+3.9)} \\ \hline
\end{tabular}}
\smallskip
\caption{Impact of targeted (TG) and non-targeted (NTG) attacks. All attacks are performed with strength $\epsilon=1$. Det-AdvProp with non-targeted attack works better on large-scale detectors and can produce more robust models against distortions. rPC denotes the relative performance under corruption.}
\vspace{-20pt}
\label{tab:tg-ntg}
\end{table}

\paragraph{Det-AdvProp with different attack strengths.}
Here we ablate the effects of attack strengths represented by the radius $\epsilon$ of the perturbation norm ball.
In Table \ref{tab:strength}, we vary $\epsilon$ from 1 to 3 and report the corresponding mAP scores on the COCO validation set.
The attacker is set as targeted since the non-targeted attack decreases the mAP of lightweight models (D0-D2) even when $\epsilon$ is small as mentioned above.
Aligning with the findings in image classification~\cite{xie2020advprop}, large perturbation size degrades the performance of small models.
EfficientDets D0-D2 work the best with $\epsilon=1$, and stronger attack can cause performance degradation.
On the contrary, EfficientDets D3-D5 work the best with relatively large perturbation $\epsilon=2$.
The clean performance of EfficientDet-D5 is boosted by 0.2 mAP by increasing the $\epsilon$ from 1 to 2.
It is reasonable to conclude that stronger attack strengths are desired for Det-AdvProp to better boost the detectors with higher capacities.

\begin{table}[!htb]
\vspace{-3pt}
\centering
\resizebox{.26\textwidth}{!}{
\begin{tabular}{c|c|c|c|c|c|c}
\hline
$\epsilon$ & D0 & D1 & D2 & D3 & D4 & D5 \\ \hline \hline
0 & 34.3 & 40.2 & 43.5 & 46.8 & 49.3 & 51.3 \\ \hline
1 & \textbf{34.7} & \textbf{40.5} & \textbf{43.8} & \textbf{47.2} & 49.6 & 51.5 \\ \hline
2 & 34.2 & 40.0 & 43.5 & \textbf{47.2} & \textbf{49.7} & \textbf{51.7} \\ \hline
3 & 34.1 & 40.0 & 43.4 & 47.1 & 49.5 & 51.6 \\ \hline
\end{tabular}}
\smallskip
\caption{Impact of attack strengths. All attacks are conducted in the targeted manner. $\epsilon=0$ means vanilla training. Larger attack strengths work better with larger model capacities.}
\vspace{-20pt}
\label{tab:strength}
\end{table}

\paragraph{Variants of Det-AdvProp.}
Different from image classifiers, object detectors need to identify objects by their class labels and box coordinates. They take the form of an inherent multi-task learning, preventing a direct application of AdvProp to the detectors. A straightforward idea would be generating adversarial examples by maximizing the total training losses $L_{det}$.
Another approach is to generate two adversarial examples per clean image based on $L_{cls}$ and $L_{loc}$ separately, assuming the adversarial images sourced from classification and localization have distinct distributions. 
We call the last method \textbf{3BN} since it constructs three batchnorms during training.
Besides, we also ablate the methods that attack the detectors based on an individual loss (either $L_{cls}$ or $L_{loc}$).
We choose non-targeted attack for this ablation study.
According to the comparison in Table \ref{tab:taskloss}, we have the following observations:

\begin{compactitem}
    \vspace{0.2em}
    \item When attacking an individual task loss, choosing classification or localization does not make a big difference on clean COCO images. 
    However, attacking $L_{cls}$ performs much better than attacking $L_{loc}$  on COCO-C, implying that the performance degradation caused by corruptions may mainly come from the classification branch. 
    When attacking the total loss $L_{det}$, the resulting detectors' performance is in between, verifying that the adversarial gradients sourced from classification and localization may mutually conflict~\cite{zhang2019towards}.
    \vspace{0.2em}
    \item \textbf{3BN} explicitly augments the detectors with both classification and localization branches, leading to the highest relative performance on corrupted images. However, the detectors fail to achieve high mAP on clean images probably because the adversarial features act as overly strong regularization. Indeed, the two auxiliary batchnorms in 3BN may dominate the optimization procedure, making the detectors under-fitting the clean training images.
    \vspace{0.2em}
    \item The proposed Det-AdvProp is the best method among all variants by consistently achieving the highest mAP on both COCO and COCO-C for the detectors of various scales.
    It uses only one auxiliary batchnorm during training to prevent excessive regularization and separately attacks the two branches to avoid the misalignment between adversarial examples.
\end{compactitem}


\begin{table}[!htb]
\vspace{-2pt}
\centering
\resizebox{.38\textwidth}{!}{
\begin{tabular}{l|c|c|c}
\hline
Model & \tabincell{c}{COCO \\ mAP} & \tabincell{c}{COCO-C \\ mAP} & \tabincell{c}{Relative \\ rPC (\%)} \\
\hline\hline
\textbf{EfficientDet-D3} & 46.8 & 28.8 & 61.5 \\ 
+ Det-AdvProp (\textbf{LOC}) & 47.1 (+0.3) & 30.0 (+1.2) & 63.7 (+2.2) \\
+ Det-AdvProp (\textbf{CLS}) & 47.2 (+0.4) & 30.5 (+1.7) & 64.6 (+3.1) \\
+ Det-AdvProp (\textbf{DET}) & 47.1 (+0.3) & 30.4 (+1.6) & 64.5 (+3.0) \\
+ Det-AdvProp (\textbf{3BN}) & 46.7 (-0.1) & 30.6 (+1.8) & \textbf{65.5 (+4.0)} \\
+ Det-AdvProp & \textbf{47.6 (+0.8)} & \textbf{30.8 (+2.0)} & 64.7 (+3.2) \\ \hline

\textbf{EfficientDet-D4} & 49.3 & 30.1 & 61.1 \\
+ Det-AdvProp (\textbf{LOC}) & 49.6 (+0.3) & 31.7 (+1.6) & 63.9 (+2.8) \\
+ Det-AdvProp (\textbf{CLS}) & 49.6 (+0.3) & 32.6 (+2.5) & 65.7 (+4.6) \\
+ Det-AdvProp (\textbf{DET}) & 49.6 (+0.3) & 32.7 (+2.6) & 65.9 (+4.8) \\
+ Det-AdvProp (\textbf{3BN}) & 49.2 (-0.1) & 32.5 (+2.4) & \textbf{66.1 (+5.0)} \\
+ Det-AdvProp & \textbf{49.8 (+0.5)} & \textbf{32.8 (+2.8)} & 65.9 (+4.8) \\ \hline

\textbf{EfficientDet-D5} & 51.3 & 31.4 & 61.2 \\
+ Det-AdvProp (\textbf{LOC}) & 51.6 (+0.3) & 33.1 (+1.7) & 64.1 (+2.9) \\
+ Det-AdvProp (\textbf{CLS}) & 51.7 (+0.4) & 33.6 (+2.2) & 65.0 (+3.8) \\
+ Det-AdvProp (\textbf{DET}) & 51.6 (+0.3) & 33.4 (+2.0) & 64.7 (+3.5) \\
+ Det-AdvProp (\textbf{3BN}) & 51.3 (+0.0) & 33.5 (+2.1) & \textbf{65.3 (+4.1)} \\
+ Det-AdvProp & \textbf{51.8 (+0.5)} & \textbf{33.7 (+2.3)} & 65.1 (+3.9) \\ \hline
\end{tabular}}
\smallskip
\caption{Comparison of several variants of Det-AdvProp. \textbf{LOC}, \textbf{CLS}, and \textbf{DET} generate the adversarial images based on $L_{loc}$, $L_{cls}$, and $L_{det}$ respectively. 
\textbf{3BN} denotes the variant that generates two adversarial examples per clean image and employs three batchnorms during training.
Det-AdvProp achieves the largest performance gains on both clean and corrupted images.
}
\vspace{-15pt}
\label{tab:taskloss}
\end{table}

\vspace{-4pt}
\paragraph{RetinaNet results.}
Apart from the state-of-the-art EfficientDets, we also test Det-AdvProp on the RetinaNet object detector~\cite{lin2017focal} with a ResNet50 backbone~\cite{he2016resnet}. 
We associate Det-AdvProp with non-targeted attack and the attack strength of $\epsilon=1$. 
The other settings are the same as the baseline method.
Det-AdvProp improves the mAP of RetinaNet from 35.6 to 36.1 on COCO and from 17.8 to 19.7 on COCO-C.
We anticipate bigger improvements if we base RetinaNets on the backbones of higher capacities. 

\vspace{-4pt}
\section{Conclusion}
In this paper, we systematically examine the data augmentation strategies for object detectors.
We discover that the performance gains on ImageNet classification including both accuracy and robustness, cannot be preserved after the object detection fine-tuning process.
Instead, the proposed Det-AdvProp is specifically crafted for the fine-tuning process.
Det-AdvProp dynamically learns from the stronger attack emerged from the classification and localization domains, which enables its policy to evolve during fine-tuning.
This model-and-data-dependent manner is more effective than previous model-agnostic augmentation strategies.
Extensive experiments show that our methods can consistently and substantially outperform the vanilla training and AutoAugment under various settings.
The obtained detector is not only more accurate, but also more robust to image distortions and domain shift.

{\footnotesize
{\noindent {\bf Acknowledgement}: We would like to thank Yingwei Li for valuable discussions. CX is supported by a gift grant from Open Philanthropy. CJH is supported in part by Army Research Laboratory under agreement number 	
W911NF-20-2-0158, and by NSF under  IIS-2008173, IIS-2048280. }
}

{\small
\bibliographystyle{ieee_fullname}
\bibliography{egbib}

\begin{thebibliography}{10}\itemsep=-1pt

\bibitem{tide-eccv2020}
Daniel Bolya, Sean Foley, James Hays, and Judy Hoffman.
\newblock Tide: A general toolbox for identifying object detection errors.
\newblock In {\em ECCV}, 2020.

\bibitem{chen2019shapeshifter}
Shang-Tse Chen, Cory Cornelius, Jason Martin, and Duen~Horng Chau.
\newblock Shapeshifter: Robust physical adversarial attack on faster r-cnn
  object detector.
\newblock {\em Lecture Notes in Computer Science}, page 52–68, 2019.

\bibitem{cubuk2019autoaug}
Ekin~D Cubuk, Barret Zoph, Dandelion Mane, Vijay Vasudevan, and Quoc~V Le.
\newblock Autoaugment: Learning augmentation strategies from data.
\newblock In {\em Computer Vision and Pattern Recognition}, 2019.

\bibitem{cubuk2020randaug}
Ekin~D Cubuk, Barret Zoph, Jonathon Shlens, and Quoc~V Le.
\newblock Practical automated data augmentation with a reduced search space.
\newblock In {\em Advances in Neural Information Processing Systems}, 2020.

\bibitem{deng2009imagenet}
Jia Deng, Wei Dong, Richard Socher, Li-Jia Li, Kai Li, and Li Fei-Fei.
\newblock Imagenet: A large-scale hierarchical image database.
\newblock In {\em Computer Vision and Pattern Recognition}, 2009.

\bibitem{pascal-voc-2012}
Mark Everingham, Luc Van~Gool, Christopher~KI Williams, John Winn, and Andrew
  Zisserman.
\newblock {The Pascal Visual Object Classes (VOC) Challenge}.
\newblock {\em International Journal of Computer Vision}, 2010.

\bibitem{kevin2018phyattack}
Kevin Eykholt, Ivan Evtimov, Earlence Fernandes, Bo Li, Amir Rahmati, Florian
  Tramer, Atul Prakash, Tadayoshi Kohno, and Dawn Song.
\newblock Physical adversarial examples for object detectors.
\newblock In {\em USENIX Conference on Offensive Technologies}, 2018.

\bibitem{geirhos2018imagenettrained}
Robert Geirhos, Patricia Rubisch, Claudio Michaelis, Matthias Bethge, Felix~A.
  Wichmann, and Wieland Brendel.
\newblock Imagenet-trained {CNN}s are biased towards texture; increasing shape
  bias improves accuracy and robustness.
\newblock In {\em International Conference on Learning Representations}, 2019.

\bibitem{ghiasi2019nasfpn}
Golnaz Ghiasi, Tsung{-}Yi Lin, and Quoc~V. Le.
\newblock {NAS-FPN:} learning scalable feature pyramid architecture for object
  detection.
\newblock In {\em Computer Vision and Pattern Recognition}, 2019.

\bibitem{goodfellow2015fgsm}
Ian Goodfellow, Jonathon Shlens, and Christian Szegedy.
\newblock Explaining and harnessing adversarial examples.
\newblock In {\em International Conference on Learning Representations}, 2015.

\bibitem{he2019rethinking}
Kaiming He, Ross Girshick, and Piotr Doll{\'a}r.
\newblock Rethinking imagenet pre-training.
\newblock In {\em International Conference on Computer Vision}, 2019.

\bibitem{he2017mask}
Kaiming He, Georgia Gkioxari, Piotr Dollar, and Ross Girshick.
\newblock Mask r-cnn.
\newblock In {\em International Conference on Computer Vision}, 2017.

\bibitem{he2016resnet}
Kaiming He, Xiangyu Zhang, Shaoqing Ren, and Jian Sun.
\newblock Deep residual learning for image recognition.
\newblock In {\em Computer Vision and Pattern Recognition}, 2016.

\bibitem{hendrycks2018benchmarking}
Dan Hendrycks and Thomas Dietterich.
\newblock Benchmarking neural network robustness to common corruptions and
  perturbations.
\newblock In {\em International Conference on Learning Representations}, 2019.

\bibitem{kurakin2017adversarial}
Alexey Kurakin, Ian Goodfellow, and S Bengio.
\newblock Adversarial machine learning at scale.
\newblock In {\em International Conference on Learning Representations}, 2017.

\bibitem{lemley2017smart}
Joseph Lemley, Shabab Bazrafkan, and Peter Corcoran.
\newblock Smart augmentation learning an optimal data augmentation strategy.
\newblock {\em IEEE Access}, 2017.

\bibitem{li2019exploring}
Yuezun Li, Xiao Bian, Ming-Ching Chang, and Siwei Lyu.
\newblock Exploring the vulnerability of single shot module in object detectors
  via imperceptible background patches.
\newblock {\em arXiv preprint arXiv:1809.05966}, 2018.

\bibitem{li2020shapetexture}
Yingwei Li, Qihang Yu, Mingxing Tan, Jieru Mei, Peng Tang, Wei Shen, Alan
  Yuille, and Cihang Xie.
\newblock Shape-texture debiased neural network training.
\newblock {\em arXiv preprint arXiv:2010.05981}, 2020.

\bibitem{lim2019fast}
Sungbin Lim, Ildoo Kim, Taesup Kim, Chiheon Kim, and Sungwoong Kim.
\newblock Fast autoaugment.
\newblock In {\em Advances in Neural Information Processing Systems}, 2019.

\bibitem{lin2017focal}
Tsung-Yi Lin, Priya Goyal, Ross Girshick, Kaiming He, and Piotr Dollar.
\newblock Focal loss for dense object detection.
\newblock In {\em International Conference on Computer Vision}, 2017.

\bibitem{lin2015coco}
Tsung-Yi Lin, Michael Maire, Serge Belongie, James Hays, Pietro Perona, Deva
  Ramanan, Piotr Doll{\'a}r, and C~Lawrence Zitnick.
\newblock Microsoft coco: Common objects in context.
\newblock In {\em European Conference on Computer Vision}. 2014.

\bibitem{liu2016ssd}
Wei Liu, Dragomir Anguelov, Dumitru Erhan, Christian Szegedy, Scott Reed,
  Cheng-Yang Fu, and Alexander~C Berg.
\newblock Ssd: Single shot multibox detector.
\newblock In {\em European conference on computer vision}, 2016.

\bibitem{liu2019dpatch}
Xin Liu, Huanrui Yang, Ziwei Liu, Linghao Song, Hai Li, and Yiran Chen.
\newblock Dpatch: An adversarial patch attack on object detectors.
\newblock In {\em AAAI Workshop on Artificial Intelligence Safety}, 2019.

\bibitem{long2015fcn}
Jonathan Long, Evan Shelhamer, and Trevor Darrell.
\newblock Fully convolutional networks for semantic segmentation.
\newblock In {\em Computer Vision and Pattern Recognition}, 2015.

\bibitem{lu2017adversarial}
Jiajun Lu, Hussein Sibai, and Evan Fabry.
\newblock Adversarial examples that fool detectors.
\newblock {\em arXiv preprint arXiv:1712.02494}, 2017.

\bibitem{madry2018towards}
Aleksander Madry, Aleksandar Makelov, Ludwig Schmidt, Dimitris Tsipras, and
  Adrian Vladu.
\newblock Towards deep learning models resistant to adversarial attacks.
\newblock In {\em International Conference on Learning Representations}, 2018.

\bibitem{michaelis2020benchmarking}
Claudio Michaelis, Benjamin Mitzkus, Robert Geirhos, Evgenia Rusak, Oliver
  Bringmann, Alexander~S Ecker, Matthias Bethge, and Wieland Brendel.
\newblock Benchmarking robustness in object detection: Autonomous driving when
  winter is coming.
\newblock {\em arXiv preprint arXiv:1907.07484}, 2019.

\bibitem{peng2018megdet}
Chao Peng, Tete Xiao, Zeming Li, Yuning Jiang, Xiangyu Zhang, Kai Jia, Gang Yu,
  and Jian Sun.
\newblock Megdet: A large mini-batch object detector.
\newblock In {\em Computer Vision and Pattern Recognition}, 2018.

\bibitem{qi2018frustum}
Charles~R Qi, Wei Liu, Chenxia Wu, Hao Su, and Leonidas~J Guibas.
\newblock Frustum pointnets for 3d object detection from rgb-d data.
\newblock In {\em Computer Vision and Pattern Recognition}, 2018.

\bibitem{redmon2016yolo}
Joseph Redmon, Santosh Divvala, Ross Girshick, and Ali Farhadi.
\newblock You only look once: Unified, real-time object detection.
\newblock In {\em Computer Vision and Pattern Recognition}, 2016.

\bibitem{ren2015faster}
Shaoqing Ren, Kaiming He, Ross Girshick, and Jian Sun.
\newblock Faster r-cnn: Towards real-time object detection with region proposal
  networks.
\newblock In {\em Advances in Neural Information Processing Systems}, 2017.

\bibitem{tan2019enet}
Mingxing Tan and Quoc Le.
\newblock Efficientnet: Rethinking model scaling for convolutional neural
  networks.
\newblock In {\em International Conference on Machine Learning}, 2019.

\bibitem{tan2020edet}
Mingxing Tan, Ruoming Pang, and Quoc~V. Le.
\newblock Efficientdet: Scalable and efficient object detection.
\newblock In {\em Computer Vision and Pattern Recognition}, 2020.

\bibitem{xie2020advprop}
Cihang Xie, Mingxing Tan, Boqing Gong, Jiang Wang, Alan~L. Yuille, and Quoc~V.
  Le.
\newblock Adversarial examples improve image recognition.
\newblock In {\em Computer Vision and Pattern Recognition}, 2020.

\bibitem{xie2017adversarial}
Cihang Xie, Jianyu Wang, Zhishuai Zhang, Yuyin Zhou, Lingxi Xie, and Alan
  Yuille.
\newblock Adversarial examples for semantic segmentation and object detection.
\newblock In {\em International Conference on Computer Vision}, 2017.

\bibitem{xie2020self}
Qizhe Xie, Minh-Thang Luong, Eduard Hovy, and Quoc~V Le.
\newblock Self-training with noisy student improves imagenet classification.
\newblock In {\em Computer Vision and Pattern Recognition}, 2020.

\bibitem{zhang2018mixup}
Hongyi Zhang, Moustapha Cisse, Yann~N. Dauphin, and David Lopez-Paz.
\newblock mixup: Beyond empirical risk minimization.
\newblock In {\em International Conference on Learning Representations}, 2018.

\bibitem{zhang2019towards}
Haichao Zhang and Jianyu Wang.
\newblock Towards adversarially robust object detection.
\newblock In {\em International Conference on Computer Vision}, 2019.

\bibitem{zhang2019trades}
Hongyang Zhang, Yaodong Yu, Jiantao Jiao, Eric~P Xing, Laurent~El Ghaoui, and
  Michael~I Jordan.
\newblock Theoretically principled trade-off between robustness and accuracy.
\newblock In {\em International Conference on Machine Learning}, 2019.

\bibitem{zhang2019interpreting}
Tianyuan Zhang and Zhanxing Zhu.
\newblock Interpreting adversarially trained convolutional neural networks.
\newblock In Kamalika Chaudhuri and Ruslan Salakhutdinov, editors, {\em
  Proceedings of the 36th International Conference on Machine Learning},
  volume~97 of {\em Proceedings of Machine Learning Research}, pages
  7502--7511. PMLR, 09--15 Jun 2019.

\bibitem{zhang2020adversarial}
Xinyu Zhang, Qiang Wang, Jian Zhang, and Zhao Zhong.
\newblock Adversarial autoaugment.
\newblock In {\em International Conference on Learning Representations}, 2020.

\bibitem{zhang2019bag}
Zhi Zhang, Tong He, Hang Zhang, Zhongyue Zhang, Junyuan Xie, and Mu Li.
\newblock Bag of freebies for training object detection neural networks.
\newblock {\em arXiv preprint arXiv:1902.04103}, 2019.

\bibitem{zoph2019learning}
Barret Zoph, Ekin~D Cubuk, Golnaz Ghiasi, Tsung-Yi Lin, Jonathon Shlens, and
  Quoc~V Le.
\newblock Learning data augmentation strategies for object detection.
\newblock {\em arXiv preprint arXiv:1906.11172}, 2019.

\end{thebibliography}
}

\clearpage

\appendix

\section{Det-AdvProp + AutoAugment Results}

\paragraph{On COCO-C.}
\label{app:coco-c}
We report the results of Det-AdvProp + AutoAugment~\cite{zoph2019learning} on COCO-C~\cite{michaelis2020benchmarking}.
As shown in Table~\ref{tab:ap-aa-cococ}, we can achieve more improvement on larger detectors and the largest performance gain is +3.8 mAP.

\begin{table}[!htb]
\centering
\resizebox{.38\textwidth}{!}{
\begin{tabular}{l|c|c}
\hline
Model & Vanilla & Det-AdvProp + AA \\
\hline\hline
\textbf{EfficientDet-D0} & 21.4 & 22.7 (+1.3) \\
\textbf{EfficientDet-D1} & 24.4 & 26.7 (+2.3) \\ 
\textbf{EfficientDet-D2} & 26.7 & 28.9 (+2.2) \\ 
\textbf{EfficientDet-D3} & 28.8 & 32.0 (+3.2) \\ 
\textbf{EfficientDet-D4} & 30.1 & 33.9 (+3.8) \\
\textbf{EfficientDet-D5} & 31.4 & 35.0 (+3.6) \\ \hline
\end{tabular}}
\smallskip
\caption{The comparison of Det-AdvProp + AutoAugment and the vanilla baseline on COCO-C.}
\vspace{-10pt}
\label{tab:ap-aa-cococ}
\end{table}

\paragraph{On PASCAL VOC.}
\label{app:voc}
We also transfer the detectors obtained by combining Det-AdvProp with AutoAugment to PASCAL VOC 2012~\cite{pascal-voc-2012}. 
Suggested by Table~\ref{tab:ap-aa-voc}, the combined strategy consistently outperforms the vanilla baseline and the largest improvement is +1.3 mAP.

\begin{table}[!htb]
\centering
\resizebox{.45\textwidth}{!}{
\begin{tabular}{l|c|c|c}
\hline
Model & mAP & AP50 & AP75 \\
\hline\hline
\textbf{EfficientDet-D0} & 55.6 & 77.6 & 61.4 \\
+ Det-AdvProp + AA & 56.2 (+0.6) & 78.3 (+0.7) & 62.3 (+0.9) \\ \hline
\textbf{EfficientDet-D1} & 60.8 & 82.0 & 66.7 \\
+ Det-AdvProp + AA & 61.3 (+0.5) & 82.5 (+0.5) & 67.6 (+0.9) \\ \hline
\textbf{EfficientDet-D2} & 63.3 & 83.6 & 69.3 \\
+ Det-AdvProp + AA & 63.6 (+0.3) & 84.0 (+0.4) & 70.0 (+0.7) \\ \hline
\textbf{EfficientDet-D3} & 65.7 & 85.3 & 71.8 \\
+ Det-AdvProp + AA & 66.4 (+0.7) & 85.9 (+0.6) & 72.8 (+1.0) \\ \hline
\textbf{EfficientDet-D4} & 67.0 & 86.0 & 73.0 \\
+ Det-AdvProp + AA & 67.8 (+0.8) & 87.0 (+1.0) & 74.3 (+1.3) \\ \hline
\textbf{EfficientDet-D5} & 67.4 & 86.9 & 73.8 \\
+ Det-AdvProp + AA & 68.7 (+1.3) & 88.0 (+1.1) & 75.4 (+1.6) \\ \hline
\end{tabular}}
\smallskip
\caption{The performance of the detectors trained by Det-AdvProp + AutoAugment on PASCAL VOC.}
\vspace{-10pt}
\label{tab:ap-aa-voc}
\end{table}

\section{Comparison with Adversarial Training}
\label{app:adv}
Existing adversarial training methods for detection sacrifice clean accuracy to gain adversarial robustness (e.g.,
Zhang \etal.~\cite{zhang2019towards} report a drop of 12\% mAP). We further confirm this by the adversarial training results in Table~\ref{tab:compare_adv} ($\epsilon = 1$ for a fair comparison).
In comparison, Det-AdvProp focuses on both clean accuracy and robustness (against common corruptions and domain shift). Table~\ref{tab:compare_adv} shows that we largely outperform adversarial training.

\begin{table}[!htb]
\centering
\resizebox{.38\textwidth}{!}{
\begin{tabular}{c|c|c|c|c}
\hline
\multirow{2}{*}{Method} & \multicolumn{2}{c}{COCO} & \multicolumn{2}{|c}{COCO-C} \\ \cline{2-5}
& D1 & D3 & D1 & D3 \\ \hline \hline
Vanilla & 40.2 & 46.8 & 24.4 & 28.8 \\ \hline
AutoAugment~\cite{zoph2019learning} & 40.1 & 47.0 & 25.1 & 29.6 \\ \hline
RandAugment~\cite{cubuk2020randaug} & 39.5 & 47.0 & 25.1 & 29.7 \\ \hline
Adv-Training~\cite{zhang2019towards} & 38.0 & 45.3 & 24.7 & 29.9 \\ \hline
Det-AdvProp & \textbf{40.5} & \textbf{47.6} & \textbf{25.6} & \textbf{30.8} \\ \hline
\end{tabular}}
\smallskip
\caption{Comparison (mAP) between Det-AdvProp and baselines on COCO and COCO-C.}
\vspace{-15pt}
\label{tab:compare_adv}
\end{table}

\section{Result Analysis}
\label{app:analysis}
\begin{figure*}[!htb]
\centering
\includegraphics[width=.99\textwidth]{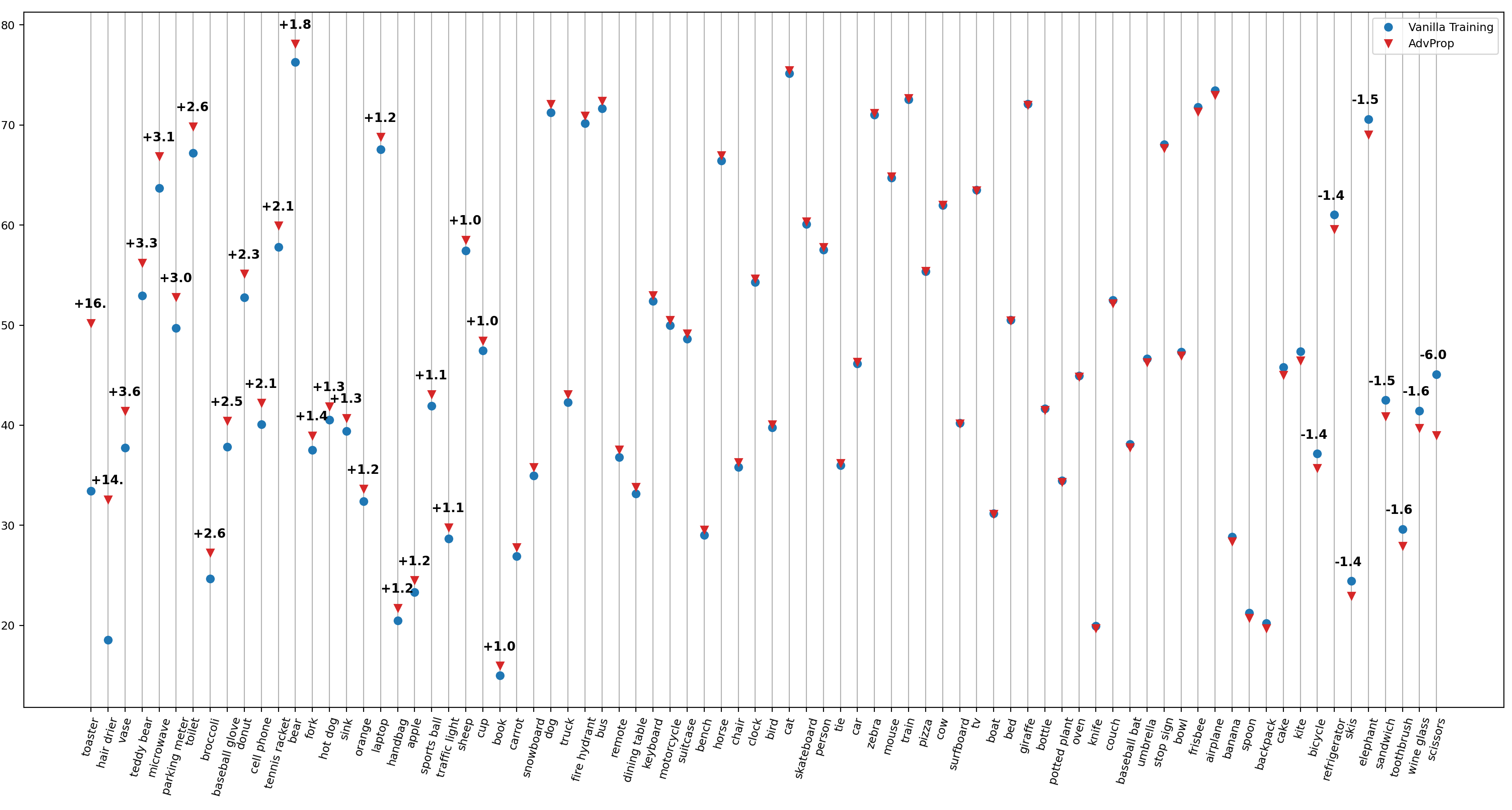}
\caption{Performance comparison of our Det-AdvProp and vanilla training on all 80 classes of COCO~\cite{lin2015coco}.
We can increase the mAP score on most classes and can even achieve +10 mAP improvement on class ``toaster'' and ``hair dryer''.}
\label{fig:class}
\end{figure*}

\paragraph{Per class results.}
We select EfficientDet-D3 trained by the proposed Det-AdvProp and show its mAP on all the 80 classes of COCO~\cite{lin2015coco}.
As illustrated in Figure~\ref{fig:class}, we can achieve improvement on most classes and can even increase the mAP score for over 10 mAP on ``toaster'' and ``hair dryer''.

\paragraph{Identify detection errors.}
We use the TIDE toolbox to segment the objection errors on COCO~\cite{lin2015coco} into six types (i.e., classification error, localization error, both classification and localization error, duplicate detection error, background error, missed ground-truth error), false positive, and false negative.
Please refer to their paper for details~\cite{tide-eccv2020}.
They identify the importance of every error type with a metric dAP, which represents the change in the overall mAP after fixing a certain error.
A large dAP for a given error means that this error type is critical, and fixing it can largely increase the overall mAP. 
As shown in Table~\ref{tab:tide}, we can reduce most error types and achieve consistent gain on \textit{classification error}, \textit{missed ground-truth error}, \textit{false positive}, and \textit{false negative} for every model.

\begin{table*}[!htb]
\centering
\resizebox{.99\textwidth}{!}{
\begin{tabular}{l|c|c|c|c|c|c||c|c}
\hline
Model & Cls & Loc & Both & Dupe & Bkg & Miss & FalsePos & FalseNeg \\
\hline\hline
\textbf{EfficientDet-D0} & 3.23 & 7.50 & 1.14 & 0.23 & 3.11 & 7.52 & 19.02 & 18.15 \\
+ Det-AdvProp + AA & \textbf{2.89 (-0.34)} & 7.81 (+0.31) & \textbf{1.12 (-0.02)} & \textbf{0.21 (-0.02)} & \textbf{3.05 (-0.06)} & \textbf{7.37 (-0.15)} & \textbf{18.34 (-0.68)} & \textbf{18.10 (-0.05)} \\ \hline
\textbf{EfficientDet-D1} & 3.00 & 7.10 & 1.09 & 0.24 & 3.50 & 5.94 & 18.30 & 15.29 \\
+ Det-AdvProp + AA & \textbf{2.85 (-0.15)} & 7.24 (+0.14) & \textbf{1.02 (-0.07)} & 0.26 (+0.02) & 3.53 (+0.03) & \textbf{5.85 (-0.09)} & \textbf{17.79 (-0.51)} & \textbf{15.21 (-0.08)} \\ \hline
\textbf{EfficientDet-D2} & 2.72 & 6.58 & 1.13 & 0.26 & 3.56 & 5.38 & 17.82 & 13.74 \\
+ Det-AdvProp + AA & \textbf{2.58 (-0.14)} & \textbf{6.54 (-0.04)} & \textbf{1.01 (-0.12)} & \textbf{0.25 (-0.01)} & 3.85 (+0.29) & \textbf{5.19 (-0.19)} & \textbf{17.52 (-0.30)} & \textbf{13.24 (-0.50)} \\ \hline
\textbf{EfficientDet-D3} & 2.44 & 6.12 & 1.14 & 0.23 & 4.12 & 4.28 & 18.09 & 11.24 \\
+ Det-AdvProp + AA & \textbf{2.42 (-0.02)} & \textbf{6.11 (-0.01)} & \textbf{1.08 (-0.06)} & 0.23 (-0.00) & \textbf{3.94 (-0.18)} & \textbf{4.08 (-0.20)} & \textbf{17.65 (-0.44)} & \textbf{10.81 (-0.43)} \\ \hline
\textbf{EfficientDet-D4} & 2.27 & 6.03 & 1.09 & 0.29 & 4.02 & 3.76 & 17.86 & 9.68 \\
+ Det-AdvProp + AA & \textbf{2.10 (-0.17)} & \textbf{5.75 (-0.28)} & 1.10 (+0.01) & \textbf{0.26 (-0.03)} & 4.20 (+0.18) & \textbf{3.68 (-0.08)} & \textbf{17.22 (-0.64)} & \textbf{9.42 (-0.26)} \\ \hline
\textbf{EfficientDet-D5} & 2.10 & 5.68 & 0.99 & 0.26 & 4.09 & 3.47 & 17.01 & 9.31 \\
+ Det-AdvProp + AA & \textbf{1.97 (-0.13)} & \textbf{5.61 (-0.07)} & 1.05 (+0.06) & \textbf{0.24 (-0.02)} & 4.26 (+0.17) & \textbf{3.19 (-0.28)} & \textbf{16.68 (-0.33)} & \textbf{8.50 (-0.81)} \\ \hline
\end{tabular}}
\smallskip
\caption{dAP (lower is better) of six detection error types, false positive, and false negative for vanilla training and our Det-AdvProp + AutoAugment. The number is bold if we successfully reduce the dAP.}
\label{tab:tide}
\end{table*}

\section{Detailed Experiment Settings}
\label{app:settings}
\paragraph{Model and training details.}
Following \cite{tan2020edet}, we train all models  using a SGD optimizer with momentum 0.9 and weight decay 4e-5.
We train every model for 300 epochs (unless specified otherwise) with batch size 256, where the learning rate increases from 0 to 0.32 in the first epoch and then cosine-decays for the rest epochs.
Same as the vanilla training of EfficientDets, we only employ horizontal flipping and a jitter uniformly sampled from $[0,2.0]$ as additional data augmentation.
We use 32 TPUv3 cores for D0-D2 and 128 TPUv3 cores for D3-D5, but the batch size for batchnorm is always set to 256 by using the synchronized batch normalization~\cite{peng2018megdet} with epsilon 1e-3 and decay 0.99.
The fast gradient sign method (FGSM) with random initialization~\cite{goodfellow2015fgsm} is chosen as our default attacker for its efficiency.
The multi-step attack performs almost the same as a single step FGSM but incurs more computation, probably because $\epsilon$ in this paper is smaller (up to 3/255) than that in standard adversarial attack (8/255).
The criterion for selecting the (targeted vs. non-targeted) attack is the clean accuracy by default unless specifically clarified, leading to targeted attack for D0---D2 (although non-targeted attack can bring better robustness shown in Table~\ref{tab:tg-ntg}) and non-targeted attack for D3---D5.
We always use $\epsilon=1$ except for the study of attack strength in Table~\ref{tab:strength}.
We also scale the input image to the range of $[-1, 1]$ for the convenience of performing attack.

\paragraph{Datasets and evaluation metrics.}
We evaluate both accuracy and robustness of our object detectors. The COCO 2017 object detection~\cite{lin2015coco} provides 118K training images for the detectors and 5K validation images for testing the detectors' within-dataset performance. 
To test the generalization ability of the trained models under various corruptions, we follow \cite{michaelis2020benchmarking} to generate the COCO-C datasets with distorted validation images.
There are 15 types of corruptions (e.g., Gaussian noise, snow, JPEG compression) each with 5 severity level (larger severity means stronger corruption strength), and we compute the mAP on COCO-C by averaging over the 75 corrupted datasets.
The relative performance under corruption (rPC) is also reported, measuring the relative performance degradation compared to the detectors' mAP on the clean images.
Moreover, we test the models' transferibility to PASCAL VOC 2012~\cite{pascal-voc-2012} dataset, which contains 20 object classes.

\paragraph{Training Complexity.}
Det-AdvProp needs additional forward and backward propagation to obtain the adversarial examples, its training complexity is about 3x than that of vanilla training. So we also compare with 3x schedule of vanilla training, which trains the detector for 900 epochs. 
With 3x schedule, the mAP of D1/D3 increases from 40.2/46.8 to 40.3/47.0 (vs.\ 40.5/47.6 by Det-AdvProp) on COCO. 
There is no change on COCO-C or VOC.
So we can still largely outperform the baseline with similar budget.

\section{Other Things We Tried}
Recent work shows that distorting the pixels within bounding boxes helps the training of object detectors~\cite{zoph2019learning}, but it is unclear whether adversarial distortion would work.
We also attempt to treat the background and object pixels differently when applying Det-AdvProp.
To achieve this, we use different attack strengths for the background and object pixels, expecting the detector to be discerning towards important features while remaining robust.
Empirical results demonstrate that applying a larger $\epsilon$ to the object pixels can slightly boost the detectors' robustness on COCO-C, but does not make a big different on COCO.
We will further explore this direction in future work.

\end{document}